\documentclass[10pt,twocolumn,letterpaper]{article}

\usepackage{cvpr}
\usepackage{times}
\usepackage{epsfig}
\usepackage{graphicx}
\usepackage{amsmath}
\usepackage{amssymb}
\usepackage[T1]{fontenc}
\usepackage{booktabs}
\usepackage{caption}
\usepackage{layouts}
\usepackage{paralist}

\usepackage[breaklinks=true,bookmarks=false,colorlinks=true]{hyperref}
\usepackage{multirow,enumitem}

\usepackage[table,x11names,dvipsnames,table]{xcolor}
\definecolor{rowblue}{RGB}{220,230,240}
\definecolor{myorchid}{RGB}{150,10,30}
\definecolor{myblue}{RGB}{10,30,250}
\definecolor{mygreen}{RGB}{10,120,10}
\definecolor{myorange}{RGB}{180,100,10}
\definecolor{fco}{RGB}{236,112,20}
\newcommand{\CR}[1]{{#1}}

\cvprfinalcopy

\newcommand{\notation}[1]{\ensuremath{#1}\xspace}

\newcommand{\Pixel}{\notation{n}}
\newcommand{\PixelD}{\notation{\delta}}
\newcommand{\Time}{\notation{t}}
\newcommand{\BasisIdx}{\notation{b}}

\newcommand{\NumFrames}{\notation{T}}
\newcommand{\NumBasis}{\notation{B}}
\newcommand{\KSize}{\notation{K}}

\newcommand{\NoiseFreeFrameTensor}{\notation{R}}
\newcommand{\NoisyFrameTensor}{\notation{I}}
\newcommand{\DenoisedFrameTensor}{\notation{\hat{\NoiseFreeFrameTensor}}}
\newcommand{\KernelTensor}{\notation{w}}
\newcommand{\BasisTensor}{\notation{v}}
\newcommand{\CoeffTensor}{\notation{c}}

\newcommand{\Conv}[2]{\notation{#1 \star #2}}

\newcommand\blfootnote[1]{%
  \begingroup
  \renewcommand\thefootnote{}\footnote{#1}%
  \addtocounter{footnote}{-1}%
  \endgroup
}

\newcommand{\NoisyFrame}[2]{\notation{\NoisyFrameTensor[#1,#2]}}
\newcommand{\NoiseFreeFrame}[2]{\notation{\NoiseFreeFrameTensor[#1,#2]}}
\newcommand{\DenoisedFrame}[1]{\notation{\DenoisedFrameTensor[#1]}}
\newcommand{\Kernel}[3]{\notation{\KernelTensor_{#1}[#2,#3]}}
\newcommand{\Basis}[3]{\notation{\BasisTensor_{#2}[#1,#3]}}
\newcommand{\Coeff}[2]{\notation{\CoeffTensor_{#1}[#2]}}
\newcommand{\IndexedConv}[3]{\notation{\left(\Conv{#1}{#2}\right)[#3]}}

\begin{document}

\title{Basis Prediction Networks for Effective Burst Denoising with Large Kernels}
\author{Zhihao Xia$^1$, Federico Perazzi$^2$, Micha\"el Gharbi$^2$, Kalyan Sunkavalli$^2$, Ayan Chakrabarti$^1$\\
  $^1$Washington University in St. Louis~~~~~~$^2$Adobe Research\\
  {\tt\small \{zhihao.xia,ayan\}@wustl.edu, \{perazzi,mgharbi,sunkaval\}@adobe.com}
}

\maketitle
\blfootnote{Part of this work was done when ZX was an intern at Adobe Research.}
  
\begin{abstract}
Bursts of images exhibit significant self-similarity across both time and space. This motivates a representation of the kernels as linear combinations of a small set of basis elements. To this end, we introduce a novel basis prediction network that, given an input burst, predicts a set of global basis kernels --- shared within the image --- and the corresponding mixing coefficients --- which are specific to individual pixels. Compared to state-of-the-art techniques that output a large tensor of per-pixel spatiotemporal kernels, our formulation substantially reduces the dimensionality of the network output. This allows us to effectively exploit comparatively larger denoising kernels, achieving both significant quality improvements (over 1dB PSNR) and faster run-times over state-of-the-art methods.
\end{abstract}

\section{Introduction}
\label{sec:intro}
\begin{figure}
   \centering
   \includegraphics[width=\linewidth]{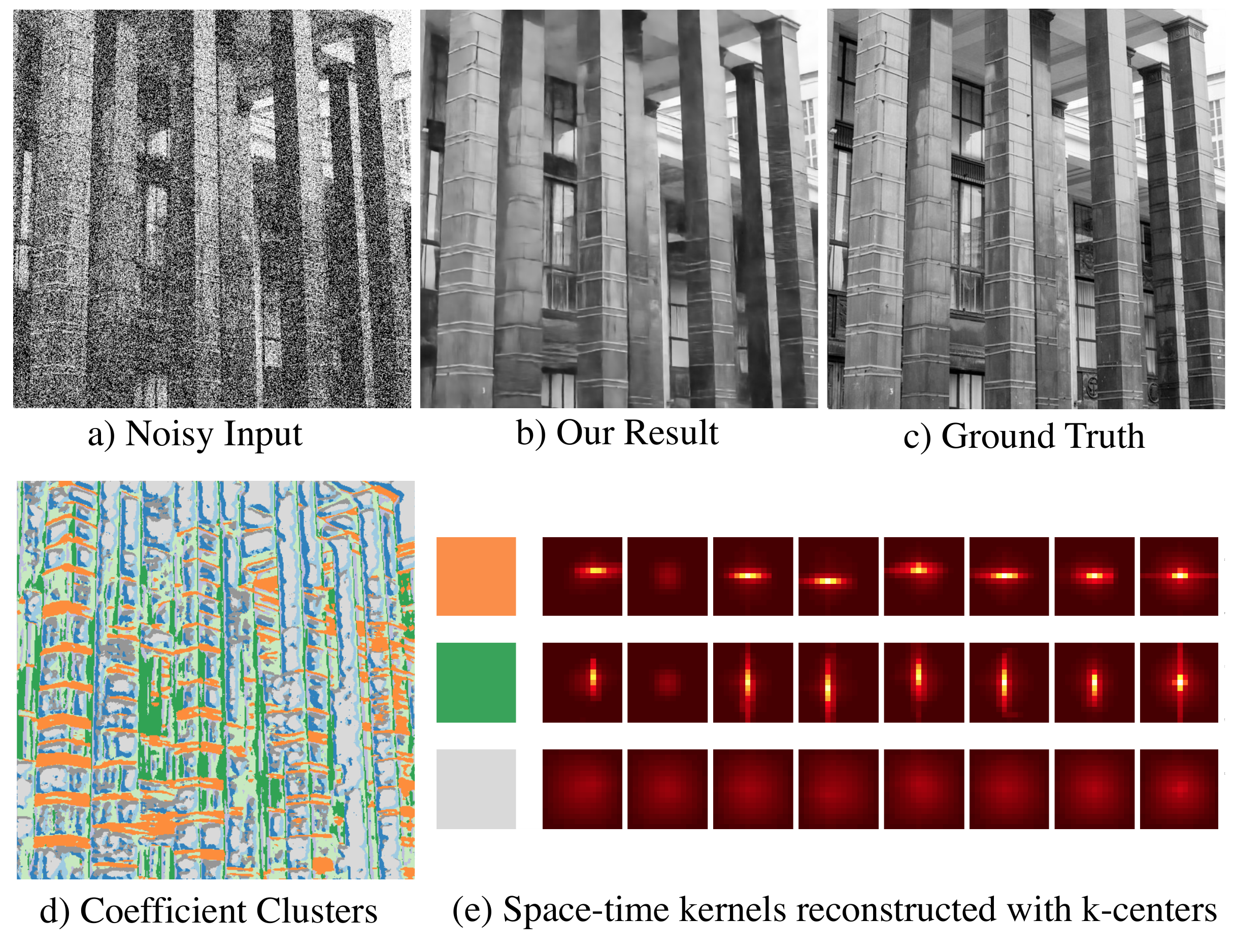}
    \caption{\label{fig: kmeans} (\emph{top}) Qualitative result of our
       denoising network. The proposed approach (b) recovers fine geometric
       details from a very noisy image burst (a, only one frame is shown).
       (\emph{Bottom}) our per-burst bases can compactly represent a large
       variety of kernels by exploiting the redundancy of local image
       structures. We clustered the per-pixel coefficients predicted from the
       noisy burst using $k$-means. The clusters we obtain show strong spatial
       structure (d). For instance, we can identify a cluster corresponding to
       horizontal image edges (e, orange), one  corresponding to vertical edges
       (e, green), and another for homogeneous regions with no structure (e,
       gray).}
\end{figure}

Burst denoising algorithms~\cite{liu2014fast,hdrplus,burstkpn} seek to
enable high-quality photography in challenging conditions and are
increasingly being deployed in commercial mobile
cameras~\cite{hdrplus,liba2019lowlight}. A burst captures a sequence
of short-exposure frames of the scene that are free of motion-blur,
but with a high amount of noise in each frame and relative motion
between frames. By accounting for this relative motion and using the
fact that the noise is independent across frames, burst denoising
attempts to aggregate these inputs and predict a single noise- and
blur-free image estimate.

Recently, Mildenhall~\etal~\cite{burstkpn} proposed an elegantly
simple yet surprisingly successful approach to burst denoising. Rather
than explicitly estimating inter-frame motion~\cite{hdrplus,
  liu2014fast, heide2016proximal, flexisp,kokkinos2019iterative},
their method produces denoised estimates at each pixel as a weighted
average of observed noisy intensities in a window around that pixel's
location in all frames. These averaging weights, or kernels, are
allowed to vary from pixel-to-pixel to implicitly account for motion
and image discontinuities, and are predicted from the noisy input
burst using a ``kernel prediction network'' (KPN).

However, KPNs need to produce an output that is significantly
higher-dimensional than the denoised image---even for $5\times 5$
kernels with eight frames, a KPN must predict 400 times as many kernel
weights as image intensities. This comes with significant memory and
computational costs, as well as difficulty in training given the many
degrees of freedom in the output. As a result, KPNs have so far been
used only with small kernels. This limits their denoising ability by
preventing averaging across bigger spatial regions, and over frames
with larger relative motion.

In this paper, we introduce an approach to predict large denoising
kernels, and thus benefit from wider aggregation, while simultaneously
limiting output dimensionality at each pixel, making the prediction
network easier to train and compute- and memory-efficient. This
approach is motivated by a long history of successful image
restoration methods that have leveraged internal structure and
self-similarity in natural
images~\cite{bm3d,nlmeans,xia2018identifying,liu2018non,zhang2019residual}. In
burst denoising, self-similarity is particularly strong because we
expect both \emph{spatial} structure in the form of similar patterns
that recur within a frame and across frames of the same scene, and
\emph{temporal} structure caused by consistency in scene and camera
motion. Given the expected self-similarity and structure in the image
intensities themselves, we argue that the corresponding denoising
kernels must also have similar structure. Specifically, while allowing
for individual per-pixel denoising kernels to be large, we assume that
all the kernels for a given image span a lower-dimensional subspace.

Based on this observation, we introduce a new method for kernel-based burst denoising that achieves better accuracy and efficiency. Our contributions are:

\begin{itemize}[nosep,leftmargin=\labelwidth]
\item \CR{We introduce basis prediction networks (BPNs): given an input noisy burst, these networks predict a global low-dimensional basis set for large denoising kernels}, and per-pixel coefficient vectors relative to this basis. BPN outputs are thus significantly lower-dimensional than the per-pixel arbitrary kernels from a regular KPN.
\item Enforcing this structure on the denoising kernels acts as a form
  of regularization, and our experiments show this leads to
  state-of-the-art denoising performance with significantly
  higher quality (>1 dB PSNR) than KPNs.
\item Beyond reducing memory usage and computational burden at the
  output layer, the structure of BPN's output enables the
  final kernel filtering step to be performed much more efficiently in
  the Fourier domain. We show this in terms of required number of
  FLOPs, as well as experimentally with actual run-times.
\end{itemize}

\section{Related Work}
\label{sec:rw}

\paragraph{Single-image and video denoising.}
Single-image denoising has been studied extensively. To overcome the
ill-posed nature of the problem, classical
approaches~\cite{foe,foe0,epll} developed regularization schemes that
model the local statistics of natural images. The most successful
approaches~\cite{bm3d, nlmeans} exploit non-local self-similarity
within the image and denoise pixels by aggregating similar pixels or
patches from distant regions. These methods have been extended to
denoise videos~\cite{vbm3d, vbm4d}, where the search for similar
patches proceeds not only within frames, but also across different
frames. Recent work has improved image denoising performance using
convolutional networks trained on large
datasets~\cite{mlp,dncnn,ircnn,liu2018non,zhang2019residual,xia2018identifying,xie2012denoising}.
Other works use hand-crafted features to select per-pixel filters from learned filter sets for efficient image enhancement~\cite{raisr,blade}.

\paragraph{Burst denoising.}
Single-image denoising is a fundamentally under-constrained problem.
Burst processing can reduce this ambiguity by using multiple
observations (the frames of the burst) to recover a noise-free
depiction of the scene.  Burst denoising algorithms are now
extensively used in commercial smartphone cameras~\cite{hdrplus} and
can produce compelling results even in extreme low light
scenarios~\cite{chen2018dark,liba2019lowlight}.  Like in video
denoising, a significant challenge for burst processing is the
robustness to inter-frame motion.  Many methods explicitly estimate
this motion to align and denoise the frames~\cite{hdrplus,
  liu2014fast,heide2016proximal, flexisp, kokkinos2019iterative}.
Current state-of-the-art burst denoising
techniques~\cite{burstkpn,kokkinos2019iterative,dburst,mkpn} are based
on deep neural networks. Many of them only require coarse
registration, relying on the network to account for the small residual
misalignments~\cite{burstkpn,dburst,mkpn}.

\paragraph{Kernel Prediction Networks.}
Given a burst sequence, Mildenhall et al.~\cite{burstkpn}
predict per-pixel kernels that are then applied to the input burst
to produce the denoised output. They demonstrate that KPNs outperform
direct pixel synthesis that produces oversmooth
results. This idea has been extended to use multiple
varying size kernels at every pixel~\cite{mkpn}. KPNs have also
been used in other applications, including denoising Monte Carlo
renderings~\cite{mckpn,mckpn2,smckpn}, video
super-resolution~\cite{sreskpn} and
deblurring~\cite{zhou2019deblurring}, frame
interpolation~\cite{videokpn,videosep,liu2017video} and video
prediction~\cite{jia2016dynamic,finn2016unsupervised,liu2017video,xue2016visual}.

Given their high-dimensional output (per-pixel kernels, that are
three-dimensional in the case of burst denoising), KPNs have
significant memory and compute requirements.
Marin{\v{c}}\etal~\cite{mkpn} and Niklaus \etal~\cite{videosep} ameliorate this by predicting
spatially separable kernels: forming spatial kernels as outer products of predicted horizontal and vertical
kernels. However, this makes a strong a-priori
assumption about kernel structure, and still requires constructing and
filtering with different per-pixel kernels. In contrast, our approach
assumes that the set of per-pixel kernels for a scene span a
low-dimensional sub-space, and predicts a basis for this sub-space
based on the burst input. This approach also allows us to benefit from
fast filtering in the Fourier domain.

\section{Method}
\label{sec:method}

\begin{figure*}[!t]
  \centering
  \includegraphics[width=\linewidth]{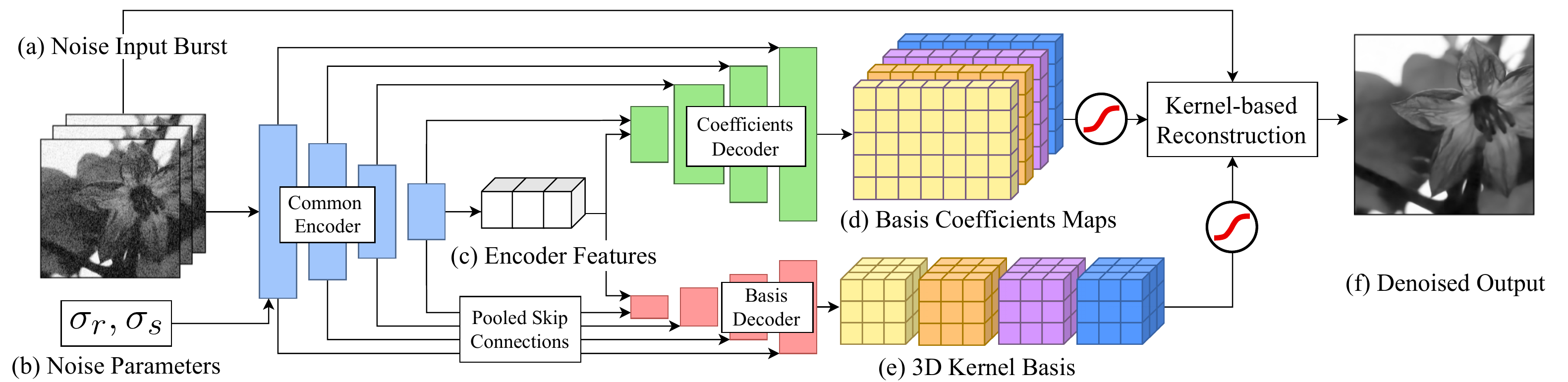}
  \caption{\label{fig:network}Our basis prediction network takes as
    input a burst of noisy input frames (a) together with the noise parameters (b).
    The frames are encoded into a shared feature space (c). These features are
    then decoded by two decoders with skip connections into a burst-specific
    basis of 3D kernels (e) and a set of per-pixel mixing coefficients (d). Both
    the coefficients and basis kernels are individually unit-normalized.
    Finally, we obtain per-pixel kernels by mixing the basis elements according
    to the coefficients and we apply them to the input burst to produce the final
    denoised image (f).
  }
\end{figure*}

In burst denoising, we are given an input noisy burst of images
$\NoisyFrame{\Pixel}{\Time}$, where
\Pixel indexes spatial locations and $\Time \in \{1,\ldots,\NumFrames\}$ the different
frames in the burst. Using a heteroscedastic Gaussian noise
model~\cite{foi2008practical}, which accounts for both read and shot noise, we
relate this to the corresponding noise-free frames $\NoiseFreeFrame{\Pixel}{\Time}$ as:
\begin{equation}
  \label{eq:nzmodel}
  \NoisyFrame{\Pixel}{\Time} \sim \mathcal{N}(\NoiseFreeFrame{\Pixel}{\Time},
  \sigma_r^2 + \sigma_s^2 \NoiseFreeFrame{\Pixel}{\Time}),
\end{equation}
where $\sigma_r^2$ and $\sigma_s^2$ are the read- and shot-noise parameters.
Choosing the first frame as reference, our goal is to produce a single denoised
image $\DenoisedFrame{\Pixel}$ as an estimate of the first noise-free frame
$\NoiseFreeFrame{\Pixel}{1}$.

\subsection{Kernel-based burst denoising}

Rather than train a network to regress \DenoisedFrameTensor directly, Kernel
Prediction Networks output a field of denoising kernels
$\Kernel{\Pixel}{\PixelD}{\Time}$, one for each pixel \Pixel at each frame \Time.
The kernels have a spatial support $\KSize\times \KSize$, indexed by $\PixelD$,
with separate weights for each frame. Given these predicted kernels, the
denoised estimate \DenoisedFrameTensor is formed as:
\begin{equation}
  \label{eq:kpn}
  \DenoisedFrame{\Pixel} = \sum_{\Time} \sum_{\PixelD}
  \Kernel{\Pixel}{\PixelD}{\Time}\NoisyFrame{\Pixel-\PixelD}{\Time}.
\end{equation}
A key bottleneck in this pipeline is the prediction of this dense kernel field
\KernelTensor, which requires producing $\KSize^2\NumFrames$ numbers at
\emph{every} pixel of the output.
Since networks with high-dimensional outputs are both expensive and require learning a large number of parameters in their last layer, KPNs have typically been used only
with small kernels ($\KSize=5$ in \cite{burstkpn}).

\subsection{Basis Prediction Networks}

Instead of directly predicting unconstrained kernels for each spatial location, we
designed a network that outputs:
\begin{inparaenum}[(1)]
\item a global \emph{kernel basis} $\Basis{\PixelD}{\BasisIdx}{\Time}$, of size $\KSize^2\NumFrames\times \NumBasis$ with $\BasisIdx 
  \in \{1,\ldots \NumBasis\}$; and
\item a \NumBasis dimensional coefficient vector $\Coeff{\Pixel}{\BasisIdx}$ at each spatial location.
\end{inparaenum}
\begin{equation}
  \label{eq:kbasis}
  \Kernel{\Pixel}{\PixelD}{\Time} = \sum_{\BasisIdx}
  \Basis{\PixelD}{\BasisIdx}{\Time} \Coeff{\Pixel}{\BasisIdx}.
\end{equation}
Note that we typically choose the number of basis kernels
$\NumBasis \ll \KSize^2\NumFrames$.  This implies that all the kernels
for a given burst lie in a low-dimensional subspace, but this subspace
will be different for different bursts, \ie the basis is
burst-specific.
This procedure allows us to recreate a full kernel field with far
fewer predictions. Assuming a $W\times H$ resolution image, we need
only make $WHB+\KSize^2TB$ predictions to effectively recreate a
kernel field of size $WHK^2\NumFrames$.

We designed our network following an encoder-decoder architecture with
skip connections~\cite{unet}. Our model, however, has two decoder
branches, one for the basis, the other for the coefficients
(Figure~\ref{fig:network}).
The encoder is shared between the two branches because the meaning of
the coefficients \CoeffTensor is dependent on the predicted basis
\BasisTensor in Equation~\eqref{eq:kbasis}, so the two outputs need to
be co-ordinated.
This encoder takes the noisy burst and noisy parameters as input, and
through multiple levels of downsampling and global average pooling at
the end, yields a single global feature vector as its encoding of the
image.
The per-pixel coefficients \CoeffTensor are then decoded from the
encoder bottleneck to the full image resolution $W\times H$, with
\NumBasis channels as output.
The common basis \BasisTensor is decoded up to \emph{distinct} spatial
dimensions --- that of the kernels $\KSize\times \KSize$ --- with
$\NumBasis \times \NumFrames$ output channels.

Since the basis branch decodes to a different spatial resolution, we need a
careful treatment of the skip connections. Unlike a usual U-Net, the encoder and
decoder feature size do not match.
Specifically, a pixel $\PixelD$ in the basis kernel
$\Basis{\PixelD}{b}{\cdot}$ has no meaningful relation to a pixel
$\Pixel$ in the input frames $\NoisyFrame{\Pixel}{\cdot}$.
Therefore, in the skip connections from the shared encoder to the
basis decoder, we apply a global spatial average pooling of the
encoder's activations, and replicate the average vector to the
resolution of the decoder layer.
This mechanism ensures the encoder information is globally aggregated
without creating nonsensical correspondences between kernel and image
locations, while allowing features at multiple scales of the encoder
to inform the basis decoder.

We ensure each of the reconstructed kernel \KernelTensor has positive
weights that sum to one, to represent averaging. We implement this
constraint using soft-max normalizations on \emph{both} the
coefficient and basis decoder outputs.
So every 3D kernel of the basis $\Basis{\cdot}{\BasisIdx}{\cdot}$ and
every coefficient vector $\Coeff{\Pixel}{\cdot}$ is normalized
individually. A more detailed description of the architecture is
provided in the supplementary.


Our network is trained with respect to the quality of the final
denoised output $\hat{R}$---with an $L2$ loss on intensities and $L1$
loss on gradients. Like \cite{burstkpn}, we additionally use a
per-frame loss to bias the network away from relying only on the
reference frame. We do this with separate losses on denoised estimates
from each individual frame of the input burst (formed as
$\hat{R}_t[n]= T\sum_\delta w_n[\delta,t]I[n-\delta,t]$). These are
added to main training loss, with a weight that is decayed across
training iterations.

\subsection{Efficient Fourier domain filtering}
Filtering by convolution with large kernels is commonly implemented in the
Fourier domain, where the filtering complexity is quasilinear in image size,
while the complexity of direct convolution scales with the product of image- and
kernel-size. But because the kernels \KernelTensor in KPNs vary
spatially, Equation~\eqref{eq:kpn} does not represent a standard convolution, ruling out this acceleration.

In our case, because our kernels are defined with respect to a small
set of ``global'' basis vectors, we can leverage Fourier-domain
convolution to speed up filtering.  We achieve this by combining and
re-writing the expressions in Eq.~\eqref{eq:kpn} and
Eq.~\eqref{eq:kbasis} as:
\begin{eqnarray}
  \label{eq:kfourier}
  \DenoisedFrame{\Pixel} &=& \sum_{\Time} \sum_{\PixelD} \Kernel{\Pixel}{\PixelD}{\Time} \NoisyFrame{\Pixel-\PixelD}{\Time}\notag\\
    &=& \sum_t \sum_{\PixelD} \sum_{\BasisIdx} \Basis{\PixelD}{\BasisIdx}{\Time}
    \Coeff{\Pixel}{\BasisIdx} \NoisyFrame{\Pixel-\PixelD}{\Time}\notag\\
    &=& \sum_b \Coeff{\Pixel}{\BasisIdx} \sum_t \sum_{\PixelD}
    \Basis{\PixelD}{\BasisIdx}{\Time} \NoisyFrame{\Pixel-\PixelD}{\Time}\notag\\
    &=& \sum_b \Coeff{\Pixel}{\BasisIdx} \sum_t
    \IndexedConv{\NoisyFrame{\cdot}{\Time}}{\Basis{\cdot}{\BasisIdx}{\Time}}{\Pixel},
\end{eqnarray}
where $\star$ denotes standard spatial 2D convolution with a spatially-uniform kernel.

In other words, we first form a set of \NumBasis filtered versions of the input burst
\NoisyFrameTensor by
standard convolution with each of the basis kernels---convolving each frame in
the burst \NoisyFrameTensor with the corresponding ``slice'' of the basis kernel---and then
taking a spatially-varying linear combination of the filtered intensities at
each pixel based on the coefficients \CoeffTensor. We can carry out these standard
convolutions in the Fourier domain as:
\begin{equation}
  \label{eq:kfr2}
  \Conv{\NoisyFrame{\cdot}{\Time}}{\Basis{\cdot}{\BasisIdx}{\Time}} =
  \mathcal{F}^{-1}\left(\mathcal{F}(\NoisyFrame{\cdot}{\Time}) \cdot 
    \mathcal{F}(\Basis{\cdot}{\BasisIdx}{\Time}) \right),
\end{equation}
where $\mathcal{F}(\cdot)$ and $\mathcal{F}^{-1}(\cdot)$ are spatial forward and
inverse Fourier transforms. This is significantly more efficient for larger
kernels, especially since we need not repeat forward Fourier transform of
the inputs \NoisyFrameTensor for different basis kernels.

\section{Experiments}
\label{sec:exp}

\CR{We closely follow Mildenhall~et~al.~\cite{burstkpn} for training and evaluation.}
Our model is designed for bursts of $\NumFrames=8$ frames with resolution $128\times 128$. Following the procedure of \cite{burstkpn}, \CR{we use} training and validation sets constructed from the Open Images dataset~\cite{openimages}, \CR{with shot and read noise parameters uniformly sampled in the log-domain: $\log(\sigma_r)\in[-3, -1.5]$, and $\log(\sigma_s)\in [-4, -2]$.} We also use \cite{burstkpn}'s set of 73 \CR{grayscale} test images for evaluation. 

Our default configuration uses bases with $B=90$ kernels of size $K=15$. We train our network (as well as all ablation baselines) using Adam~\cite{kingma2014adam} with a batch-size of 24 images, and an initial learning rate of $10^{-4}$. We train for a total of about 600k iterations, dropping the learning twice, by $\sqrt{10}$ each time, whenever the validation loss saturates.

\begin{table}[!t]\small
  \centering
   \setlength\tabcolsep{4pt}
  \begin{tabular}{lcccc}
    \toprule
    Method & Gain $\propto$ 1  & Gain $\propto$ 2& Gain $\propto$ 4  & Gain $\propto$ 8 \\
    \midrule
    HDR+~\cite{hdrplus} & 31.96 & 28.25 & 24.25 & 20.05 \\
    BM3D~\cite{bm3d} & 33.89 & 31.17 & 28.53 & 25.92 \\
    NLM~\cite{nlmeans} & 33.23 & 30.46 & 27.43 & 23.86 \\
    VBM4D~\cite{vbm4d} & 34.60 & 31.89 & 29.20 & 26.52 \\
    Direct & 35.93 & 33.36 & 30.70 & 27.97 \\
    \addlinespace
    KPN*~\cite{burstkpn} & 36.47 & 33.93 & 31.19 & 27.97 \\
    KPN ($\KSize=5$) & 36.35 & 33.69 & 31.02 & 28.16 \\
    MKPN~\cite{mkpn}& 36.88 & 34.22 & 31.45 & 28.52 \\
    \addlinespace
    \CR{BPN (ours)} & \textbf{38.18} & \textbf{35.42} & \textbf{32.54} & \textbf{29.45} \\
    \bottomrule
  \end{tabular}
  \caption{\label{tab: trans} Denoising performance on a synthetic \CR{grayscale} benchmark~\cite{burstkpn}. We report performance in terms of Average PSNR
    (dB). As in \cite{burstkpn}, our method was not trained on the
    noise levels implied by the largest gain (fourth column). KPN and MKPN refer to our implementation of these techniques, while numbers for KPN* correspond to those reported in the KPN paper itself~\cite{burstkpn}. Results for all other methods\CR{, including end-to-end regression to denoised intensities (denoted as Direct)}, are from \cite{burstkpn}. Our method
    widely outperforms all prior methods at all noise levels.}
\end{table}

\begin{figure*}
   \centering
   \includegraphics[width=0.89\linewidth]{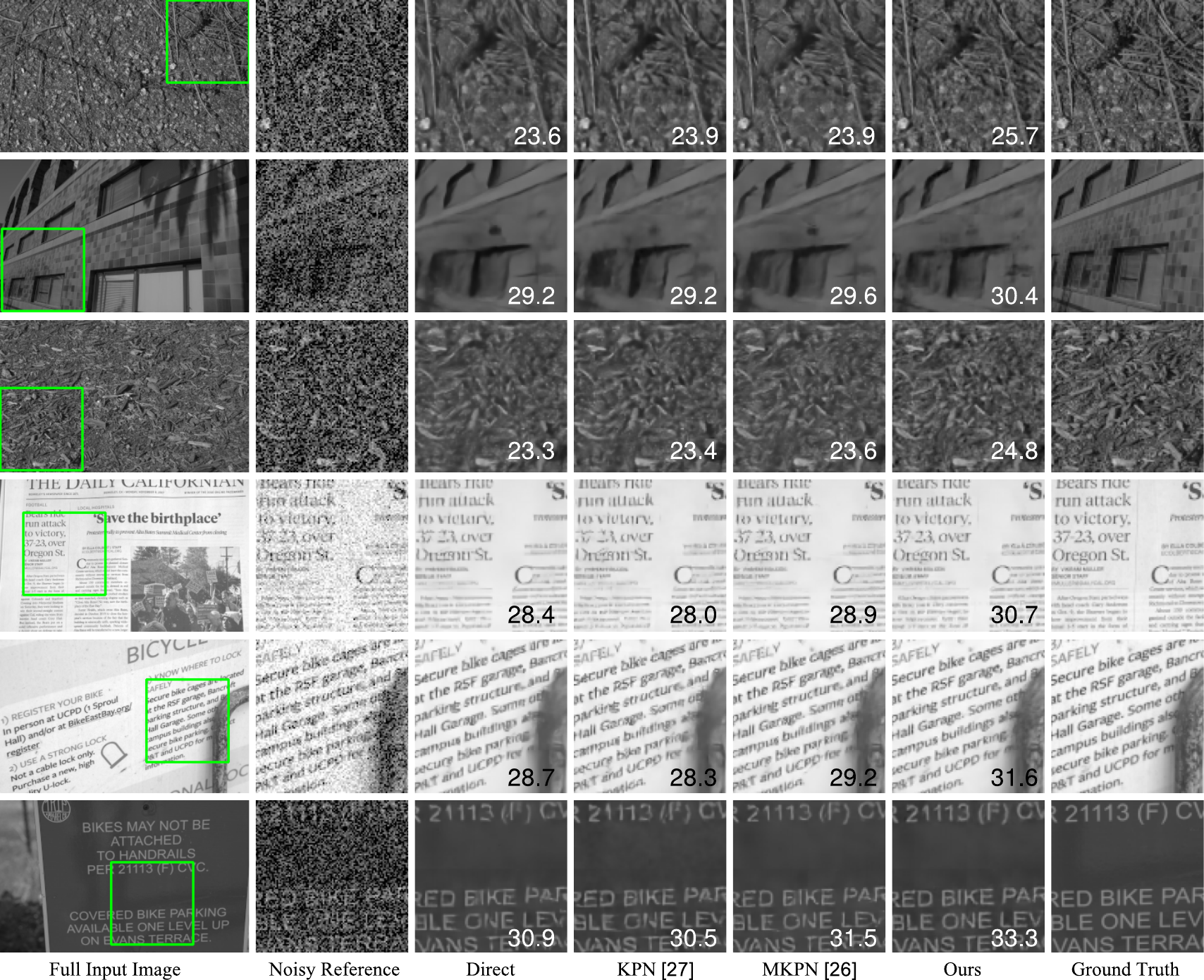}
    \caption{We illustrate denoising performance on a benchmark
       synthetic \CR{grayscale} test set~\cite{burstkpn} for our method, a direct prediction
       network (which directly regresses denoised pixels), and two KPN
       variants~\cite{burstkpn,mkpn} with the same kernel size $\KSize=15$ as
       our method. The numbers in inset refer to the PSNR (dB) on the
       full image. In addition to better quantitative performance, our method
       does better at reproducing perceptual details like textures, edges, and
       text. }
    \label{fig: result} 
\end{figure*}

A reference implementation of our method is available at \href{https://www.cse.wustl.edu/~zhihao.xia/bpn/}{https://www.cse.wustl.edu/\textasciitilde{}zhihao.xia/bpn/}.

\subsection{Denoising performance}

Table~\ref{tab: trans} reports the PSNR of our denoised outputs on the \CR{grayscale} test set~\cite{burstkpn}. Each noise level corresponds to a sensor gain value (one stop increments of the ISO setting in a camera). \CR{The gains correspond to the following values for $(\log(\sigma_s), \log(\sigma_r))$:  $1\rightarrow(-2.2, -2.6), 2\rightarrow(-1.8, -2.2), 4\rightarrow(-1.4, -1.8), 8\rightarrow(-1.1, -1.5)$.} The highest noise level, denoted as $\text{Gain} \propto 8$, lies outside the range we trained on. We use it to evaluate our model's extrapolation capability. In addition to our own model, we also report results for a motion-alignment-based method~\cite{hdrplus}, several approaches based on non-local filtering~\cite{nlmeans,bm3d,vbm4d}, as well as the standard KPN burst denoiser~\cite{burstkpn}---which is the current state-of-the-art. Since we did not have access to the original KPN model, we implemented a version ourselves (that we use in ablations in the next section) and also report its performance in Table~\ref{tab: trans}. We find it closely matches the performance reported in \cite{burstkpn}. Additionally, we train a network to directly regress the denoised pixel values from the input burst (\ie, without kernels), as well as our implementation of \cite{mkpn} with a larger kernel size of $K=15$ for fair comparison. 

We find that our method outperforms KPN~\cite{burstkpn} by a significant margin, over 1 dB PSNR at all noise levels. Our implementation of \cite{mkpn} also does well, but remains inferior to our model. We show qualitative results for a subset of methods in Figure~\ref{fig: result}. Our have fewer artifacts, especially in textured regions and around thin structures like printed text.

\subsection{Ablation and analysis}
\begin{table}[!t]\small
  \centering
  \setlength\tabcolsep{0.75pt}
  \begin{tabular}{lcccc}
    \toprule
    & Gain $\propto$ 1  & Gain $\propto$ 2& Gain $\propto$ 4  & Gain $\propto$ 8 \\
    \midrule
    KPN ($\KSize=15$) & 34.29 & 31.80 & 28.23 & 24.86 \\
    Separable ($\KSize=15$) & 34.67 & 32.05 & 28.52 & 25.12 \\
    \addlinespace
    Ours ($\KSize=5$) & 35.70 & 33.02 & 29.16 & 25.57 \\
    Ours ($\KSize=9$) & 36.22 & 33.41 & 29.56 & 25.94 \\
    \addlinespace
    Ours ($\NumBasis= 10$) & 35.31 & 32.69 & 28.95 & 25.43 \\
    Ours ($\NumBasis= 50$) & 36.10 & 33.33 & 29.45 & 25.88 \\
    Ours ($\NumBasis= 130$) & 36.27 & 33.47 & 29.57 & \textbf{25.99} \\
    \addlinespace
    \textbf{Ours ($\mathbf{\KSize=15}$, $\mathbf{\NumBasis = 90}$)} & \textbf{36.29} & \textbf{33.57} & \textbf{29.62} & \textbf{25.99} \\
    \addlinespace
    Common Spatial Basis & 35.71 & 33.04 & 29.23 & 25.71 \\
    Per-frame Spatial Basis & 36.21 & 33.46 & 29.56 & 25.92 \\
    \addlinespace
    Fixed basis & 34.66 & 32.15 & 28.68 & 25.39 \\
    \bottomrule
  \end{tabular}
  \caption{\label{tab:ablation} Ablation study on our validation dataset. Performance is reported in terms of Average PSNR (dB).
  Beyond motivating our parameter choices ($\KSize=15$,$\NumBasis = 90$), this demonstrates that our use of a burst-specific spatio-temporal basis outperforms standard KPN~\cite{burstkpn}, separable spatial kernels, a common spatial basis for all burst frames, separate spatial bases per-frame, and a fixed, input-agnostic basis. All these variants were trained with the same settings ($\KSize=15$, $\NumBasis = 90$) as our model.}
\end{table}

Our approach leads to better denoising quality because it enables larger kernels without vastly increasing the network's output dimensionality and number of learnable parameters. To tease apart the contributions of kernel size and the structure of our kernel decomposition, we conduct an ablation study on our validation set. The results can be found in Table~\ref{tab:ablation}.
The performance gap between the test and validation set results (Table~\ref{tab: trans} and~\ref{tab:ablation}) comes from differences in the datasets themselves.

\noindent\textbf{Kernel Size.} As a baseline, we consider using KPN directly with our larger kernel size of $K=15$. We also consider predicting a single separable kernel at that size (\cite{mkpn} predicts separable kernels at multiple sizes, and adds them together). We find that our network outperforms the large kernel KPN variant at all noise levels---suggesting that simply increasing the kernel size is not enough. It also outperforms separable kernel prediction, suggesting that a low-dimensional subspace constraint better captures the structure of natural images than spatial separability.

For completeness, we also evaluate our basis prediction network with smaller kernels, $K=9$ and $K=5$. Although, this leads to a drop in performance compared to our default configuration, these variants still perform better than the original KPN---suggesting our approach has a regularizing effect that benefits even smaller kernels.

\noindent\textbf{Basis Size.} The number of basis elements in our default
configuration, $B=90$, was selected from a parameter search on the validation set.
We include this analysis in Table~\ref{tab:ablation}, reporting PSNR values for
$B$ ranging from 10 to 130. We find that bases with fewer than 90 kernels lead
to a drop in quality. The larger bases, $B=130$, also performs very slightly
worse than $B=90$. We hypothesize that large bases start to have too many
degrees of freedom. This increases the dimensionality of the network's output,
which negates the benefits of a subspace restriction.

\noindent\textbf{Spatial vs.\ Spatio-temporal Basis Decomposition.} Note that we define our basis as a subspace to span 3D kernels---i.e., each of our basis elements $v_b$ is a 3D spatio-temporal kernel. We predict a single weight $c_n[b]$ at each location, which is applied to corresponding spatial kernels $v_n[\cdot,t]$ for all frames $t$. However, there are other possible choices for decomposing 3D kernels, and we consider two of these in our ablation (Table~\ref{tab:ablation}). In both cases, we output coefficients $c_{n,t}[b]$ that vary per-frame, in addition to per-location---and are interpreted as separate coefficients corresponding to a spatial basis kernel. In one case, we use a \emph{common spatial basis} $v_b[\delta]$ across all frames, with $w_n[\delta,t] = \sum_b c_{n,t}[b] v_b[\delta]$. In the other, we have a \emph{per-frame spatial basis} $v_{b,t}[\delta]$ for each frame, and $w_n[\delta,t] = \sum_b c_{n,t}[b] v_{b,t}[\delta]$. The per-frame basis increases the dimensionality of our coefficient output and leads to a slight drop in performance, likely due to a reduced regularizing effect. The common spatial basis, however, suffers a greater performance drop since it also forces kernels in all frames to share the same subspace.

\begin{figure*}
   \centering
   \includegraphics{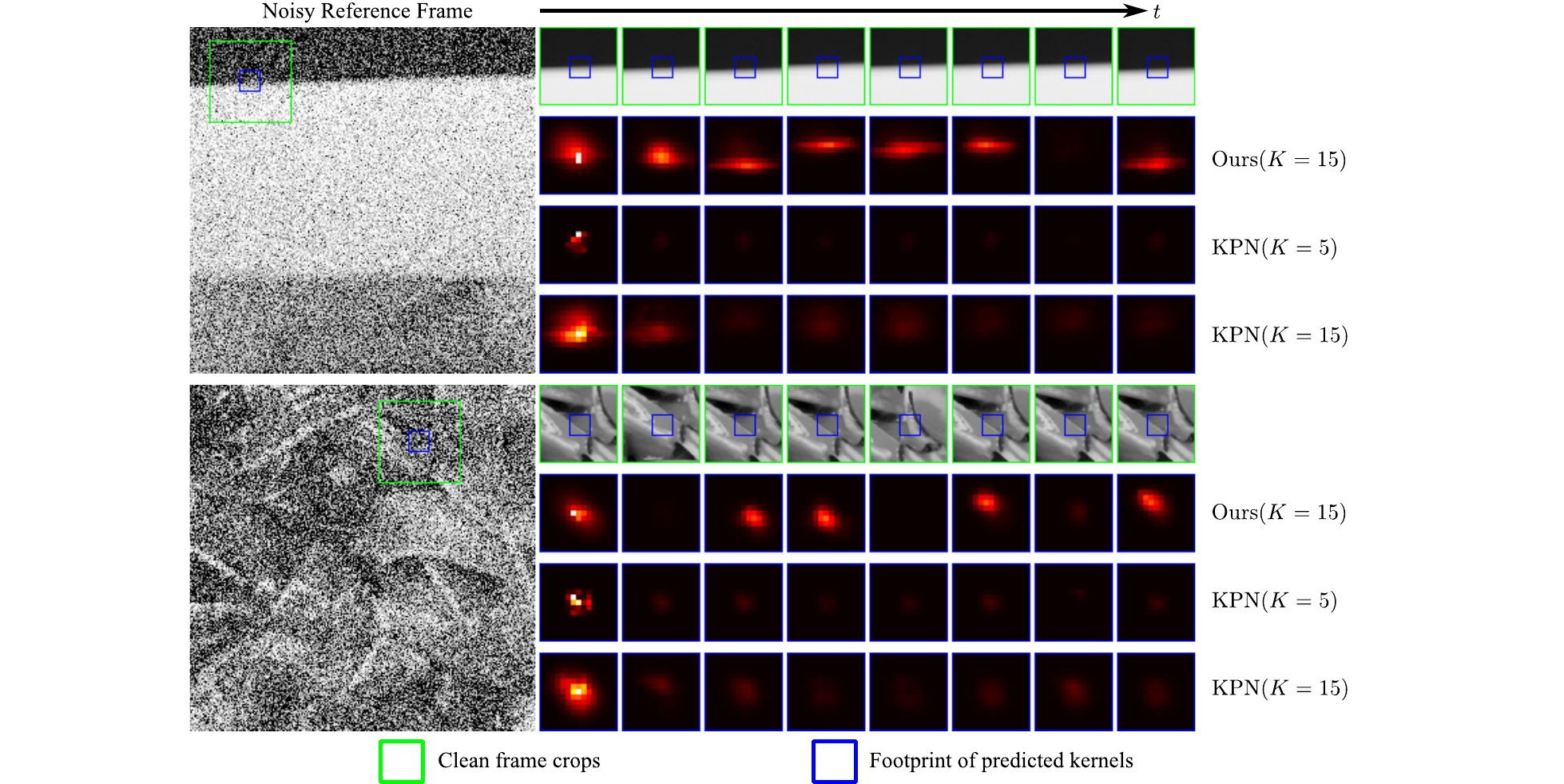}
    \caption{\label{fig: bkviz} We visualize a few 3D kernels predicted by our approach (with
       $K=15$), and those produced by standard KPN (with $K=5$ and $K=15$). For kernels
       predicted at a given location, we also show crops of the different
       noise-free frames centered at that point, with the support of the
       kernel marked in blue. In comparison to those from KPN, our
       kernels are more evenly distributed across all frames in the
       burst, with spatial patterns that closely follow the
       apparent motion in the burst.}
\end{figure*}

We also compare qualitatively the spatio-temporal kernels produced by our
default configuration with those predicted by standard KPN in Figure~\ref{fig:
bkviz}. Our model makes better use of the temporal information, applying large
weights to pixels across many frames in the burst, whereas KPN tends to overly
favor the reference frame. Our network better tracks the apparent motion in the
burst, shifting the kernel accordingly. And it is capable of ignoring outliers
caused to excessive motion (all black kernels in Fig.~\ref{fig: bkviz}).

\noindent\textbf{Fixed vs.\ Burst-specific Basis.} Given that our network predicts both a basis and per-pixel coefficients, a natural question is whether a burst-specific kernel basis is even needed. To address this, we train a network architecture without a basis decoder to only predict coefficients for each burst, and instead learn a basis that is fixed across all bursts \CR{in the training set}. The fixed basis is learned jointly with this network as a direct learnable tensor. Table~\ref{tab:ablation} shows that using a fixed basis in this manner leads to a significant decrease in denoising quality (although still better than standard KPN).

This suggests that while a subspace restriction on kernels is useful, the ideal
subspace is scene-dependent and must be predicted adaptively. We further explore
this phenomenon in Table~\ref{tab:rank}, where we quantify the rank of the
predicted bases for individual images, and for pairs of images.
Note that the rank can be lower than $B$, since we do not effectively require
the `basis' vectors $\{v_b[\cdot,\cdot]\}$ to be linearly independent.
We find that the combined rank of basis kernels of image pairs (obtained by
concatenating the two bases) is nearly twice the rank obtained from individual
images---suggesting limited overlap between the basis sets of different images.
We also explicitly compute the average overlap ratio across image pairs as $1 - rank(v, v')/[rank(v) + rank(v')]$, and find it to be around $5\%$ on average. This low overlap implies that different bursts do indeed require different bases, justifying our use of burst-specific bases.

\begin{table}[!t]\small
  \centering
  \setlength\tabcolsep{2pt}
  \begin{tabular}{lcccc}
    \toprule
     & Gain $\propto$ 1  & Gain $\propto$ 2& Gain $\propto$ 4  & Gain $\propto$ 8 \\
    \midrule
    $rank(v)$ & 80.6 & 81.8 & 84.3 & 86.2 \\
    \addlinespace
    $rank(v, v')$ & 152.2 & 154.5 & 159.6 & 165.2 \\
    Overlap ratio & 5.4\% & 5.5\% & 5.4\% & 4.4\% \\
    \bottomrule
  \end{tabular}
  \caption{\label{tab:rank} Average basis rank for each noise level (first row), average rank of the union of two bases from random burst pairs (second row), and the average overlap ratio (third row) between the subspaces spanned by the two bases. The low overlap justifies our prediction of a burst-specific basis.}
\end{table}
\begin{table}[!t]\small
  \centering
  \setlength\tabcolsep{2pt}
 \begin{tabular}{lccc}
    \toprule
    & GFLOPs & Runtime (s) \\
    \midrule
    KPN ($\KSize=15$) & 59.3 & 0.63 \\
    Separable ($\KSize=15$) & 29.9 & 0.43 \\
    \addlinespace
    Ours ($\KSize=5$) & 28.9 & 0.24\\
    Ours ($\KSize=9$) & 29.1 & 0.29\\
    \addlinespace
    Ours ($\NumBasis= 10$) & 26.5  & 0.19\\
    Ours ($\NumBasis= 50$) & 28.2 & 0.27\\
    Ours ($\NumBasis= 130$) & 31.7 & 0.41\\
    \addlinespace
    \textbf{Ours ($\mathbf{\KSize=15}$, $\mathbf{\NumBasis = 90}$)} & 29.9 & 0.30\\
    \addlinespace
    Common Spatial Basis & 40.8 & 0.49 \\
    Per-frame Spatial Basis & 41.9 & 0.57 \\
  \bottomrule
   \end{tabular}
  \caption{\label{tab:runtime} FLOPS and runtimes on 1024$\times$768 resolution
    images for different KPN denoising approaches. All variants of our basis
    prediction network are significantly faster than KPN and match the compute
    cost of separable filters (with better denoising quality). Increasing the
    kernel size for our technique comes at marginal cost thanks for the Fourier
    filtering approach. This allows us to use large kernels for better denoising
    performance.}
\end{table}

\subsection{Computational expense}

Next, we evaluate the computational expense of our approach and compare it to
the different ablation settings considered in Table~\ref{tab:ablation},
including standard KPN. We report the total number of floating point operations
(FLOPs) required for network prediction and filtering in Table~\ref{tab:runtime}. 
We find that in addition to producing higher-quality results, our approach also
requires significantly fewer FLOPs than regular KPN \CR{for the same kernel size}.
This is due to the reduced complexity of our final prediction layer, as well as
efficient filtering in the Fourier domain. Also, we find that our
approach has nearly identical complexity as separable kernel prediction, while
achieving higher denoising performance because it can express a more general
class of kernels.

In addition to the evaluation FLOPs, Table~\ref{tab:runtime} reports measured
running times for the various approaches, benchmarked on a 1024 $\times$ 768
image on an NVIDIA 1080Ti GPU.
To compute these timings, we divide the image into 128$\times$128 non-overlapping
patches to form a batch and send it to the denoising network.  
Since regular KPN has high memory requirements, we select the maximum
batch size for each method and denoise the entire image in multiple runs. This
maximizes GPU throughput.
We find that our approach retains its running time advantage
over KPN in practice. It is also a little faster than separable kernel
prediction---likely due to the improved cache performance we get from using Fourier-domain
convolutions. 

\begin{table}[!t]\small
  \centering
   \setlength\tabcolsep{4pt}
  \begin{tabular}{lcccc}
    \toprule
    Method & Gain $\propto$ 1  & Gain $\propto$ 2& Gain $\propto$ 4  & Gain $\propto$ 8 \\
    \midrule
    Direct & 38.16 &  35.39 & 32.50 & 30.27 \\
    KPN ($\KSize=5$) & 38.86 & 35.97 & 32.79 & 30.01 \\
    \addlinespace
    \CR{BPN (ours)} & \textbf{40.16} & \textbf{37.08} & \textbf{33.81} & \textbf{31.19} \\
    \bottomrule
  \end{tabular}
  \caption{\label{tab: trans_color} Denoising performance on our synthetic color test
    set. We report performance in terms of Average PSNR (dB). 
    Here, KPN refers to our extension of~\cite{burstkpn} to produce color kernels.
    Our method outperforms KPN by more than 1 dB at all noise levels.}
\end{table}

\subsection{Color burst denoising}

\CR{Finally, we report results on a color burst denoising task. We use a similar observation model as \eqref{eq:nzmodel} with noise independently added to each color channel (note this ignores multiplexed measurements and demosaicking). For kernel denoising, we use separate kernels for each color channel at each location. We extend standard KPN~\cite{burstkpn} to produce this directly, and modify our method to have the basis decoder produce a ``color'' kernel basis (of size $3\KSize^2\NumFrames\times \NumBasis$), while the coefficient decoder still outputs a \NumBasis dimensional coefficient vector.}

\CR{We use the same training protocol as for grayscale images, using color versions of the Open Images dataset. In this case, training takes 1900k iterations with batches of 8 color images. We construct a new synthetic test set of 100 images from the Open Images validation dataset, with no overlap with our training set. We report comparisons in Table~\ref{tab: trans_color}, showing a similar improvement over KPN~\cite{burstkpn} as for grayscale images---over 1 dB PSNR at all noise levels. We include qualitative comparisons in Figure~\ref{fig: color}.}

\newcommand{\includecolor}[1]{
    \includegraphics[height=2.45cm]{color_results/#1_gt_full.png} &
    \includegraphics[height=2.45cm]{color_results/#1_nz_crop.png} &
    \includegraphics[height=2.45cm]{color_results/#1_direct_crop.png} &
    \includegraphics[height=2.45cm]{color_results/#1_kpn_crop.png} &
    \includegraphics[height=2.45cm]{color_results/#1_ours_crop.png} &
    \includegraphics[height=2.45cm]{color_results/#1_gt_crop.png}\vspace{-0.3em}
}

\begin{figure*}[!h]
  \centering
\setlength{\tabcolsep}{5pt}
{\small \begin{tabular}{cccccc}
Full Input Image & Noisy Reference & Direct & KPN~\cite{burstkpn} & Ours & GT \\
\includecolor{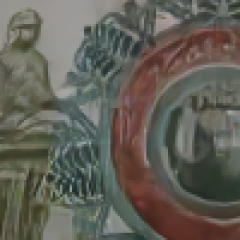} \\
& & 28.36 & 29.02 & 29.73 &\\
\includecolor{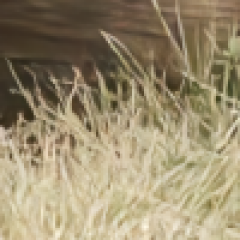} \\
& & 29.55 & 29.79 & 30.95 &\\
\end{tabular}}
\caption{\CR{We show examples of color denoising using our method on our synthetic color test set, comparing these to direct prediction and our color-extended version of KPN~\cite{burstkpn} (with $\KSize=5$). Numbers refer to PSNR (dB) on the full image.}}
\label{fig: color} 
\end{figure*}

\section{Conclusion and future work}
\label{sec:conc}

In this work, we argue that local, per-pixel burst denoising kernels are highly coherent. Based on this, we present a basis prediction network that jointly infers a global, low-dimensional kernel basis and the corresponding per-pixel mixing coefficients that can be used to construct per-pixel denoising kernels. This formulation significantly reduces memory and compute requirements compared to prior kernel-predicting burst denoising methods, allowing us to substantially improve performance by using large kernels, while reducing running time.

While this work focuses on burst denoising, KPN-based methods have been applied to other image and video enhancement tasks including video super-resolution~\cite{sreskpn}, frame interpolation~\cite{videokpn,videosep,liu2017video}, video prediction~\cite{jia2016dynamic,finn2016unsupervised}, and video deblurring~\cite{zhou2019deblurring}. All these tasks exhibit similar structure and will likely benefit from our approach.

There are other forms of spatiotemporal structure that can be explored to build on our work. For example, image enhancement methods have exploited self-similarity at different scales~\cite{glasner2009sr} suggesting other decompositions in scale space. Also, we assume a fixed basis size globally. Adapting this spatially to local content could yield further benefits. Finally, KPNs are still, at their heart, local filtering methods and it would be interesting to extend our work to non-local filtering methods~\cite{bm3d,vbm3d,vbm4d}.

\noindent\textbf{Acknowledgments.} ZX and AC acknowledge support from NSF award IIS-1820693, and a gift from Adobe Research.

{\small
\bibliographystyle{ieee_fullname}
\bibliography{refs}
}

\clearpage

\onecolumn
\appendix

\begin{center}
  {\bf Supplementary Material\\}
\end{center}

\section{Architecture}

\begin{figure}[!b]
  \begin{center}
    \includegraphics[width=0.95\textwidth]{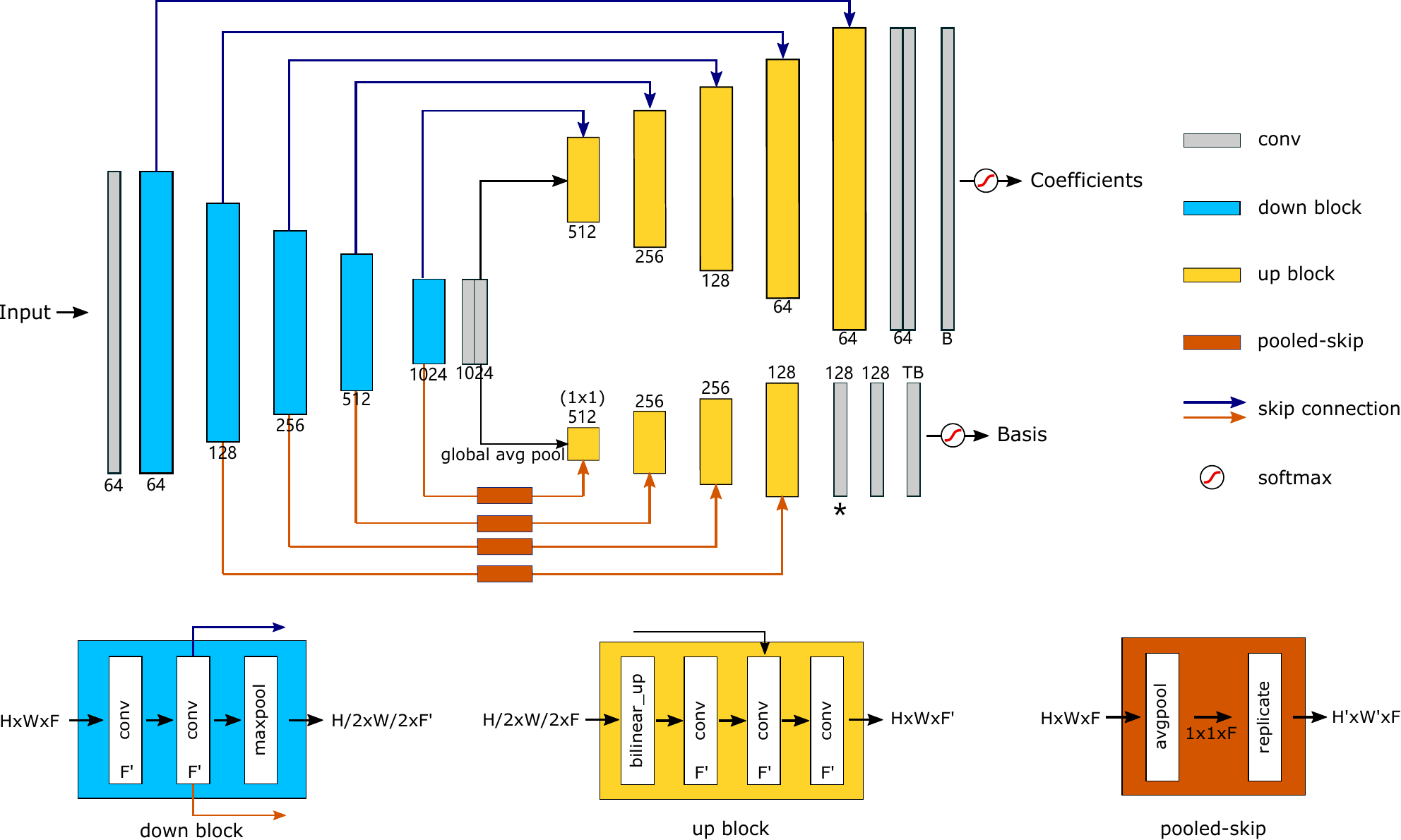}
  \end{center}
  \caption{Details for our basis prediction network. All convolutional layers are 3x3 convolutional layers with same padding, except $\star$ which is a 2x2 convolutional layer with valid padding. The down-sampling block reduces the spatial size of the input by 2 with a 2x2 stride 2 max-pooling layer. The up-sampling block has a bilinear up-sampling layer with a factor of 2. The pooled skip connection first applies a global spatial average pooling of the encoder's activations and replicate the average vector to the resolution of the decoder layer.}
  \label{fig:archi}
\end{figure}

We now provide a detailed description of the architecture of our basis prediction network, which includes a shared encoder and two decoders.  
The shared encoder network consists of five down-sampling blocks and encodes noisy frames into a shared feature space. The coefficients decoder decodes these features into a set of per-pixel mixing coefficients. Coefficients for each output pixel are normalized with a softmax function separately. 
The basis decoder first reduces these features to a 1D vector, and then decodes them to an output of shape $K\times K \times TB$, representing a burst-specific set of $B$ basis kernels, each of shape $K\times K \times T$ ($K=15$ and $B=90$ for our model).
Each 3D basis element is normalized with a softmax function so that the basis kernel sums up to 1. We include regular skip connections from the encoder to the coefficient decoder, and pooled-skip connections to the basis decoder. The entire architecture is illustrated in Figure~\ref{fig:archi}. (For color burst denoising, the decoder output has shape $K\times K\times 3TB$, representing a set of $B$ basis kernels, with each 4D basis element being $K \times K \times 3 \times T$.)

In our ablation study, we considered versions of our network that produced smaller kernels with $K=5$ and $K=9$. For $K=5$, we used one less up-sampling block in the basis decoder than for the $K=15$ case shown in Figure~\ref{fig:archi}, and added one 3$\times$3 convolutional layer with valid padding before the layer $\star$ in Figure~\ref{fig:archi}. For $K=9$, we also used one less up-sampling block in the basis decoder, but in this case, replaced the layer $\star$ in with a 2$\times$2 transpose convolution layer with valid padding (i.e., one that does a 2$\times$2 ``full'' convolution). 

While previous works~\cite{raisr,blade} learns a global set of kernel for all images and apply light-weight local processing to hard select a kernel for each pixel, our method is more powerful as it can output any kernel of larger size by predicting per-burst basis and per-pixel coefficients.

Our fixed basis ablation replaced the entire decoder branch with just a learned tensor, of size $K\times K\times (TB)$, to serve as the basis. However, we still retain the encoder and coefficient decoder, and the weights of these networks are learned jointly with the fixed basis tensor. Note that in this case, the denoising kernels at each pixel are formed as a linear combination of this fixed set of basis kernels, based on the coefficients predicted from the input burst by the decoder at each location (this is different from approaches that select one kernel at each location from a fixed kernel set~\cite{raisr,blade}). As our ablation showed, having a fixed basis set yields worse denoising performance than an adaptive basis.

\section{Additional Results}
\begin{figure*}[!h]
  \centering
\setlength{\tabcolsep}{5pt}
{\small \begin{tabular}{cccccc}
Full Input Image & Noisy Reference & Direct & KPN~\cite{burstkpn} & Ours & GT \\
\includecolor{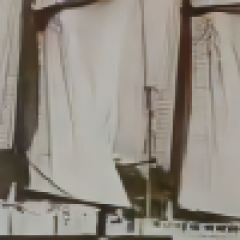} \\
& & 31.19 & 30.62 & 31.68 &\\
\includecolor{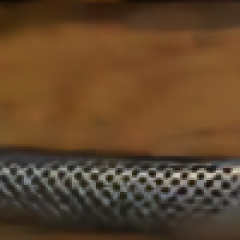} \\
& & 30.97 & 30.63 & 31.50 &\\
\includecolor{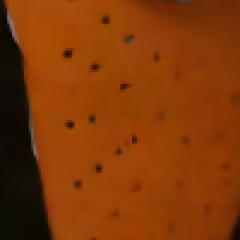} \\
& & 30.19 & 30.20 & 30.67 &\vspace{-1em}
\end{tabular}}
\caption{\CR{Additional color results on our synthetic color test set. Numbers refer to PSNR (dB) on the full image.}}
\label{fig: addcolor} 
\end{figure*}

We show additional denoising results below: for color burst denoising in Fig.~\ref{fig: addcolor} and for grayscale denoising in Figs.~\ref{fig: resultA}-\ref{fig: resultC}.

\newcommand{\includemore}[1]{
    \multicolumn{2}{c}{\includegraphics[height=3.5cm]{results/#1_nz_full.png}} &
    \multicolumn{2}{c}{\includegraphics[height=3.5cm]{results/#1_ours_full.png}} &
    \multicolumn{2}{c}{\includegraphics[height=3.5cm]{results/#1_gt_full.png}} \\[-3pt]
    \multicolumn{2}{c}{Noisy Ref.} & \multicolumn{2}{c}{Ours} & \multicolumn{2}{c}{GT} \\[2pt]

    \includegraphics[height=2.8cm]{results/#1_nz_crop.png} &
    \includegraphics[height=2.8cm]{results/#1_direct_crop.png} &
    \includegraphics[height=2.8cm]{results/#1_kpn_crop.png} &
    \includegraphics[height=2.8cm]{results/#1_mkpn_crop.png} &
    \includegraphics[height=2.8cm]{results/#1_ours_crop.png} &
    \includegraphics[height=2.8cm]{results/#1_gt_crop.png} \\[-3pt]
    Noisy Ref. & Direct & KPN~\cite{burstkpn} & MKPN~\cite{mkpn} & Ours & GT 
}

\begin{figure*}[!h]
  \centering
\setlength{\tabcolsep}{1.3pt}
\begin{tabular}{cccccc}
\includemore{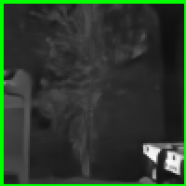} \\
 & 33.29 & 33.40 & 33.72 & 34.35 &\vspace{-0.5em}
\end{tabular}
\caption{Additional grayscale results on benchmark test set~\cite{burstkpn}. Numbers refer to PSNR (dB) on the full image.}
\label{fig: resultA} 
\end{figure*}

\begin{figure*}
   \centering
\setlength{\tabcolsep}{1.3pt}
\begin{tabular}{cccccc}

\includemore{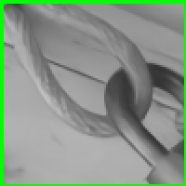} \\ 
& 34.29 & 34.27 & 34.82 & 35.52 & \\[24pt]

\includemore{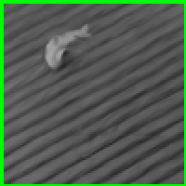} \\ 
PSNR (dB) & 28.32 & 28.90 & 29.20 & 30.34 & \\\\\\

\end{tabular}
\caption{Additional grayscale results on benchmark test set~\cite{burstkpn}. Numbers refer to PSNR (dB) on the full image.}
\end{figure*}

\begin{figure*}
   \centering
\setlength{\tabcolsep}{1.3pt}
\begin{tabular}{cccccc}
\includemore{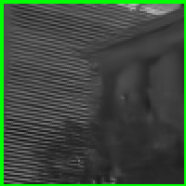} \\ 
& 29.21 & 28.64 & 29.36 & 30.35 & \\[24pt]
\includemore{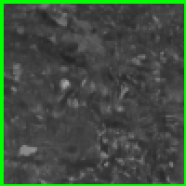} \\
&28.81 & 29.17 & 29.33 & 30.86 &  \\\\\\
\end{tabular}
\caption{Additional grayscale results on benchmark test set~\cite{burstkpn}. Numbers refer to PSNR (dB) on the full image.}
\label{fig: resultB} 
\end{figure*}

\begin{figure*}
   \centering
\setlength{\tabcolsep}{1.3pt}
\begin{tabular}{cccccc}
\includemore{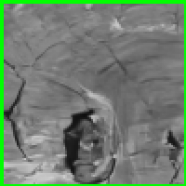} \\
 & 30.66 & 30.92 & 31.22 & 32.60 & \\[24pt]
\includemore{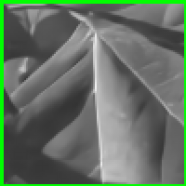} \\
 & 34.20 & 34.68 & 35.19 & 36.34 &  \\\\\\

\end{tabular}
\caption{Additional grayscale results on benchmark test set~\cite{burstkpn}. Numbers refer to PSNR (dB) on the full image.}
\label{fig: resultC} 
\end{figure*}

\end{document}